\documentclass[11pt]{article}

\usepackage{microtype} 
\usepackage{booktabs}  
\usepackage{url}  

\usepackage{amsmath}
\usepackage{amsthm}

\DeclareMathOperator*{\argmin}{arg\,min}

\usepackage[preprint]{automl}
\usepackage{algorithm}
\usepackage{algpseudocode}
\algnewcommand{\LineCommentCont}[1]{\State /* #1 */}

\usepackage{natbib}
\usepackage{todonotes}
\colorlet{lightred}{red!25}

\title{Half Search Space is All You Need}

\author[1]{\nameemail{Pavel Rumiantsev}{pavel.rumiantsev@mail.mcgill.ca}}
\author[1]{\nameemail{Mark Coates}{mark.coates@mcgill.ca}}

\affil[1,2]{The Department of Electrical and Computer Engineering\\ McGill University}
\affil[1,2]{Mila}

\hypersetup{%
  pdfauthor={Pavel Rumiantsev, Mark Coates}, 
  pdftitle={Half Search Space is All You Need},
  pdfsubject={Neural Architecture Search},
  pdfkeywords={NAS, One-Shot NAS, Zero-Shot NAS, Efficient NAS}
}

\begin{document}

\maketitle

\begin{abstract}

Neural Architecture Search~(NAS) is a powerful tool for automating architecture design.
One-Shot NAS techniques, such as DARTS, have gained substantial popularity due to their combination of search efficiency with simplicity of implementation.
By design, One-Shot methods have high GPU memory requirements during the search.
To mitigate this issue, we propose to prune the search space in an efficient automatic manner to reduce memory consumption and search time while preserving the search accuracy.
Specifically, we utilise Zero-Shot NAS to efficiently remove low-performing architectures from the search space before applying One-Shot NAS to the pruned search space.
Experimental results on the DARTS search space show that our approach reduces memory consumption by 81\% compared to the baseline One-Shot setup while achieving the same level of accuracy.

\end{abstract}


\section{Introduction}

Neural Architecture Search~(NAS) is an approach to automate the design of a neural network.
Unlike in a manual design, a researcher presents a set of all possible designs to the NAS algorithm, which searches for the best design.
It is common for NAS not only to save time and computation usage, but also to outperform manual designs.
The early approaches were characterised by a large computational overhead~\citep{elsken2019neural, zoph2018learning}.
Addressing this, \citet{liu2018darts} introduced a One-Shot NAS paradigm (identifying the optimal architecture by a single training) by treating operation selection as a determination of the weights over a fixed operation set.
All candidate operations (e.g., 3x3 convolution or maximum pooling) from a predefined set are evaluated as parallel branches.
The weighted sum is taken over the outputs of all candidate operations.
By making weights of the sum differentiable, the selection is represented as a differentiable operation.
This enables the operation selection task to be addressed as an optimisation problem within an end-to-end differentiable framework, utilising efficient gradient backpropagation, resulting in improved performance with significantly reduced search costs~\citep{kang2023neural}.

Although this design leads to a fast search with excellent performance, it suffers from excessive memory usage during the search stage as the gradients associated with all candidate operations should be backpropagated at the same time~\citep{xu2019pcdarts}.
In particular, this makes applying One-Shot NAS to large search spaces very memory demanding.
Thus, for practical application, in a setting where there are strict memory constraints, a search space has to be very carefully designed before One-Shot search is applied. The aim is to make it as small as possible via manual design~\citep{chen2019progressive}.

The alternative search paradigm is to evaluate each architecture without an expensive optimisation procedure.
Zero-Shot approaches estimate the quality of each candidate architecture using ranking functions~\citep{abdelfattah2021zerocost,mellor2021neural}.
These architecture fitness estimation approaches are fast and memory efficient compared to One-Shot NAS methods.
In particular, the Zero-Shot techniques scale much more readily to larger search spaces compared to One-Shot NAS methods.
Zero-Shot NAS approaches are generally good at identifying architectures with poor performance, but struggle to determine which of two top-performing architectures is better~\citep{mok2022demystifying,ning2021evaluating}.
Although some recent Zero-Shot approaches come close to competing with One-Shot NAS methodologies in performance~\citep{cavagnero2023freerea,li2022zico,sun2023unleashing}, One-Shot NAS is still the more popular choice due to its greater reliability when applied to a new search space.

Although Zero-Shot and One-Shot NAS complement each others' strengths and weaknesses, they cannot be used simultaneously.
However, a Zero-Shot method may be applied to significantly reduce the initial search space, pruning the architectures that are clearly poor candidates.
After pruning, One-Shot NAS can be applied to the resulting reduced search space.
The reduction of the search space reduces the memory requirements and speeds up the computation~\citep{wang2023fp}.
Therefore, this combination requires fewer resources compared to a One-Shot method applied to the initial search space and produces more accurate results than using only Zero-Shot on the initial search space.
The pipeline presented in this paper utilises both Zero- and One-Shot NAS.
It is faster and more memory efficient than One-Shot NAS, while being more precise than the original Zero-Shot NAS.

\subsection*{Contributions}

\begin{itemize}[leftmargin=*]
\item We present a general Zero-One-Shot NAS framework that is designed to easily extend the applicability of One-Shot techniques to larger search spaces. 
\item We demonstrate through experiments how our approach, when compared to One-Shot methods, significantly reduces memory and time consumption during the search phase while preserving the evaluation accuracy.

\end{itemize}

\section{Related work}

\subsection{One-Shot architecture search}

\citet{liu2018darts} proposed DARTS, the foundational idea for all One-Shot NAS techniques, by formulating the architecture search as an optimisation problem.
It introduces a weighted sum (or mixing operation) across the candidate operation set, effectively transforming the architecture search to the task of identifying the optimal weight for each operation.
The specialised search model, which is composed of multiple mixing operations, is commonly referred to as a supernetwork. 
The optimal network is derived by replacing each mixing operation with the operation that has the highest weight within its respective set.

\citet{chu2020fair} pointed out that the original design of DARTS suffers from stability and generalisation issues.
Typically, these deficiencies are addressed by adding regularisation~\citep{chu2020fair,zela2020Understanding,chu2021darts,ye2022bdarts,movahedi2023lambdadarts}. 
Such an approach increases robustness while requiring minimal change to the existing framework.
The core generalisation issue is addressed by \citet{zela2020Understanding} via $L_2$ regularisation on the parameters.
\citet{ye2022bdarts} address the stability issue by introducing beta-decay regularisation in $\beta$-DARTS. 
One of the recent works, $\Lambda$-DARTS~\citep{movahedi2023lambdadarts} tries to solve the generalisation issue by aligning gradients from different operators.
Although regularisation improves robustness, it does not address computational complexity.
Moreover, in some cases (e.g., $\Lambda$-DARTS), heavy regularisation increases complexity.

Several works directly address the computational complexity and memory requirements of DARTS.
\citet{dong2019searching} present the idea of reducing computational complexity by estimating the operator distribution through sampling.
PC-DARTS~\citep{xu2019pcdarts} improves memory efficiency by selectively processing only a randomly selected portion of the input channels of a mixing operation (subsets of the network’s internal tensor representation).

\citet{cai2023epc} introduce EPC-DARTS, an updated version of PC-DARTS that incorporates an attention module to evaluate channel importance, allowing the selection of an important subset of channels instead of random ones to enhance performance. 
In a similar effort to optimise efficiency, FP-DARTS~\citep{wang2023fp} decomposes the operator search space into two separate disconnected search spaces, each explored within individual supernetworks. 
This separation reduces the computational burden required for backpropagation.
Further improvement could be achieved by combining it with the idea of partial connection from PC-DARTS to minimise memory demands.
Furthermore, \citet{hu2024ldarts} propose a stability compensation module for PC-DARTS, tailored for the search for reduced-depth architectures, which boosts performance in stability-challenging scenarios.

In this work, we address the computational complexity issue by eliminating low-performing architectures from the search space prior to the application of One-Shot architecture search.
By filtering out these suboptimal architectures, the size of the search space is significantly reduced.
In particular, the remaining search space can still be effectively represented as a supernetwork, ensuring full compatibility with all the techniques presented above.
We provide examples to illustrate this in the experiment section.



\subsection{Zero-Shot architecture search}

Zero-Shot approaches to NAS search for the best architecture without training any neural networks. Such methods usually consist of two parts: a ranking function to estimate the fitness of an architecture, and a search algorithm to traverse the search space.
The first Zero-Shot ranking functions were described by~\citet{abdelfattah2021zerocost}. These were repurposed neural network pruning metrics; their sum over all architecture parameters can be used as an architectural fitness measure.
The first Zero-Shot approach specifically designed for NAS was presented by \citet{mellor2020neural}.
Many recent works in this field either base the ranking function on a Neural Tangent Kernel~\citep{xu2021knas,chen2021tenas,mok2022demystifying} or the ReLU activation pattern~\citep{chen2021tenas,cavagnero2023entropic}.
Some works use architecture gradients directly~\citep{li2022zico,yang2023sweet,sun2023unleashing}.
One of the fastest Zero-Shot techniques is presented by \citet{rumiantsev2023performing}; it uses architecture logits to estimate fitness without training. In our work, we use this approach.

Most Zero-Shot techniques employ a search algorithm from one of the following three families of algorithms.
(i) Random search algorithms select a random subset of candidate architectures and use the ranking function to identify the best candidates~\citep{mellor2021neural}.
(ii) Evolutionary search techniques employ genetic search procedures such as mutation and crossover. 
The procedure iterates between sampling a set of mutations of the best architectures, and estimating them using Zero-Shot ranking functions.
This allows the search to explore architectures that are similar to promising candidates. \citet{lin2021zen} present a mutation-driven evolutionary search that provides a good trade-off between performance and speed. An advanced version of this search also includes crossover and converges faster~\citep{cavagnero2023freerea}.
(iii) Pruning-based search methods~\citep{chen2021tenas} start with the supernetwork that represents all candidate architectures. The unfit operations are removed iteratively from the hypergraph.
This type of search requires evaluating the entire hypergraph for each unpruned edge, leading to considerable computational overhead.
Nevertheless, we are using this search algorithm in the current paper, since it introduces a natural connection to One-Shot NAS by utilising a supernetwork.

\section{Problem statement}

A hypergraph representation of a search space is given. 
By conducting a DARTS-based search on this hypergraph, we can assess the performance of the optimal architecture identified.
The objective is to identify an architecture whose performance meets or exceeds that of DARTS, while ensuring that the search utilises less memory in the process.

\section{Methodology}


\subsection{Masked One-Shot}
One of the main drawbacks of One-Shot is a memory requirement.
Let us consider a single mixing operation inside a DARTS cell that uses a set of trainable operations $O$, in which each element, $o(\cdot)$, is a fixed operation with trainable parameters (e.g., 3$\times$3 conv).
Each operation has its own trainable parameters that lead to high memory consumption.
To fit the One-Shot search into a GPU with a predetermined amount of memory, one must either reduce the batch size during search (reducing speed and stability~\citep{xu2019pcdarts}), or manually pruning the search space (requires manual work and possibly reduces performance~\citep{wang2023fp}).

Our approach aims to minimise the memory utilisation by pruning the search space in automatic way before applying the One-Shot NAS.
In order to preserve compatibility with the original search space, we introduce a pruning mask.
A pruned search space can be represented as a mask over the original search space.
Thus, the mixing operation with the mask is defined as
\begin{equation}
    \label{eq:masked_mixup}
    Mix(x, O) = \sum_{o \in O} \frac{ \mathbb{I}(o) \exp \left(  \alpha^o \right) }{ \sum_{o' \in O} \mathbb{I}(o') \exp \left(  \alpha^{o'} \right)  } o(x),
\end{equation}
where $O$ is a set of operations, $\mathbb{I}: O \xrightarrow{}\{ 0, 1\}$ is a pruning function that defines what operations are masked.
In practice, we never run forward and backward propagation through masked operations, therefore reducing memory consumption as well as speeding up computations.
We apply masks to the partially connected mixing operation from~\citet{xu2019pcdarts} in a similar way.

\subsection{Zero-Shot pruning}

\begin{algorithm}[t]
\caption{Partial search space pruning} \label{alg:partial_prunning}
\begin{algorithmic}[1]
\Require search hypergraph $\mathcal{N}_0$, set of ranking functions $\mathcal{R}$, pruning level $\xi \in (0,1)$

\Statex /* $\mathcal{E}_i$ is an $i$-th mixing operation set of $\mathcal{N}_t$ */
\While{$\frac{\prod_{\mathcal{E}_i \text{ in } \mathcal{N}_t} |\mathcal{E}_i|}{\prod_{\mathcal{E}_i \text{ in } \mathcal{N}_0} |\mathcal{E}_i|} > \xi$ }
    \For{edge $e_j$ in $\mathcal{N}_t$}
        \State $s_j \gets 0$ 
        \Comment {importance of operation $e_j$}
        \For{$r \in \mathcal{R}$}
            \State $s_j \gets s_j + \frac{r(\mathcal{N}_t) - r(\mathcal{N}_t \char`\\ e_j) }{r(\mathcal{N}_t)}$
        \EndFor
    \EndFor
    \State $\mathcal{N}_{t+1} \gets \mathcal{N}_t$
    \For{$\mathcal{E}_i$ in $\mathcal{N}_t$}
        \State $j \gets \argmin_j [s_j : e_j \in \mathcal{E}_i]$
        \Comment find the lowest importance operation
        \State $\mathcal{N}_{t+1} \gets \mathcal{N}_{t+1} \char`\\ e_j$ 
        \Comment prune operation $e_j$
    \EndFor

\EndWhile
\end{algorithmic}
\end{algorithm}

In order to prune the search space and generate the pruning mask, we utilise Zero-Shot NAS.
The prune-based search algorithm was initially presented by~\citet{chen2021tenas} and then generalised by~\citet{rumiantsev2023performing} for an arbitrary ranking function.
This search operates on the same hypergraph as One-Shot NAS uses, and is thus highly suitable for our method.
We base our partial pruning search~(Algorithm~\ref{alg:partial_prunning}) on a generalised prune-based search.
For clarity, we utilise the same notation and assumptions as \citet{rumiantsev2023performing}.
The substantial difference of our algorithm from the original one is a stopping condition at line 1.
As the original algorithm was designed for Zero-Shot NAS, it performs pruning until only one architecture is left.
For our needs, we apply this search partially by defining a pruning level hyperparameter $\xi \in (0,1)$.
This threshold defines the fraction of architectures that remain unpruned on the current hypergraph. We stop pruning when the fraction of architectures remaining drops below $\xi$.  

\section{Experiments}

\subsection{Dataset and experimental setup}

In order to demonstrate the efficiency of our technique, we conducted our experiments on CIFAR-10 and CIFAR-100 datasets~\citep{krizhevsky2009learning}, two widely used computer vision datasets.
Both datasets consist of 60k $32\times 32$ colour images divided equally into classes.
The datasets are split into 50k images of a training set and 10k images of a test set.
CIFAR-10 has 10 classes, each class containing 6k images.
CIFAR-100 has 100 classes, each class containing 600 images.

In our experiments, we use the DARTS search space~\citep{liu2018darts}. 
In order to ensure the consistency of the One-Shot NAS we use a PC-DARTS search setup~\citep{xu2019pcdarts} in all the experiments.
This setup is used for the search and train phases of stand-alone One-Shot models as well as masked ones.

During the search, we use the reduced search space of $8$ stacked cells ($6$ normal cells and $2$ reduction cells).
We partition the training set into two parts. The first is used to optimise each operators' trainable weights and the second is used for search parameters.
We pretrain the hypernetwork for $15$ epochs, optimising only trainable weights to alleviate the drawback of the parameterised operations.
The SGD optimiser is used for trainable weights.
We set momentum to $0.9$, weight decay to $3\times 10^{-4}$, and the initial learning rate to $0.1$.
The learning rate is annealed to $0.0$ using a cosine schedule with no restart.
The Adam optimiser~\citep{kingma2014adam} is used for search parameters.
We set a learning rate to $6\times 10^{-4}$, $\beta_1 = 0.5$, $\beta_2 = 0.999$ and a weight decay to $10^{-3}$.

After the search we evaluate the architecture.
For the evaluation, we compose a network of 20 cells (18 normal cells and 2 reduction cells).
Each cell shares the same structure obtained from the search stage.
Following the DARTS evaluation procedure~\cite{liu2018darts} the network is trained from scratch for 600 epochs with a batch size of 128.
The SGD optimiser is used to train the network.
We set momentum to $0.9$, weight decay to $3\times 10^{-4}$, the initial learning rate to $0.025$, and apply gradient norm clipping at a norm of $5$.
We apply droppath regularisation with a rate of 0.3 and a cutout~\citep{devries2017cutout}.

For the Zero-Shot NAS part of our setup we follow the ETENAS setup~\citep{rumiantsev2023performing}.
We use the Frobenius norm of the NNGP kernel as a ranking function.
This ranking function was specifically designed for fast search and reduced memory utilisation.
We estimate the NNGP kernel~\citep{park2020nngp} with 384 datapoints (three batches of size 128) for each network.
Using this ranking function we apply Algorithm~\ref{alg:partial_prunning} on the reduced search space of $8$ stacked cells.
We use a pruning level $\xi = 0.5$ in the experiments.
This setup is used for search in stand-alone Zero-Shot experiments as well as during the pruning mask search.

In order to run our experiments we use an NVIDIA Tesla V100 GPU with 32GB of VRAM.
We use eight Intel Xeon Gold 6140 CPUs with base frequency of 2.30 GHz and 16 threads each.
Our hardware setup has 64GB of RAM.

\subsection{Baselines and techniques}

Our main baselines are PC-DARTS~\citep{xu2019pcdarts} (for memory and time) and $\beta$-DARTS (for accuracy).
We provide reproduced results of DARTS~\citep{liu2018darts} and ETENAS~\citep{rumiantsev2023performing} for reference.
All baseline techniques are reproduced on our setup for consistency.
From Table~\ref{tab:memory_results}, one can see that PC-DARTS shows effective memory reduction.
As it uses only a quarter of all channels, it reduces the required memory by 70\% (it is less than the expected number due to allocations unrelated to the mixing operation).
However, it is not as efficient in reducing search time (only 14\% reduction).
We have selected ETENAS over other Zero-Shot techniques as it requires not only no training, but also no backpropagation during the search, thus reducing memory consumption by 98\% and time by 73\%.
We also make a version of it applied to a graph with partial connections (ETENAS-PC).
That further reduces the search time (by 76\% compared to DARTS), marginally increasing the memory consumption due to implementation of mixing operations with partial connections.
Memory consumption is constant for all the baselines above.
That happens because One-Shot baselines run forward and backward computations for all operations on the hypergraph at once.
Zero-Shot baselines run forward computations on the entire hypergraph at the beginning of the pruning algorithm.
Thus, memory consumption stays constant.

We present two options for pruning mask generation: Zero-Shot and random (RND in Table~\ref{tab:memory_results}).
Zero-Shot masking follows the partial pruning algorithm.
We utilise ETENAS with partial connection, since it requires less time without decreasing the evaluation accuracy.
For random masking, we generate the mask by sampling each element individually from a Bernoulli distribution with $\xi$ as a parameter.
Thus, $1-\xi$ operations are masked on average.
We ensure that the mask contains at least one unmasked operation for every set of candidate operations.
Although random masking does not consider the operation value at all, it brings close to zero computational overhead, thus reducing search time.

In addition to the masked mixing operation, our approach relies on two One-Shot techniques: partial connection from PC-DARTS and beta regularisation from $\beta$-DARTS.
Partial connection is an efficient memory reduction technique.
As it uses only a quarter of all channels, it reduces the required memory by 70\% (less than the expected amount due to allocations unrelated to the mixing operation).
However, it is not as efficient in reducing search time.
Beta regularisation is a simple technique to increase stability of the search. It does not affect search memory, and it marginally increases search time since it requires additional gradient flows to be computed.

\begin{table}[t]
    \centering
    \caption{Results on the DARTS search space. The table shows peak VRAM memory utilisation for the search stage in MB, search time in seconds and evaluation accuracy for each memory reduction technique. Baselines reproduced on our setup. Techniques with -RND suffix use randomly sampled masks.}
    \label{tab:memory_results}
    \begin{tabular}{l|l|l|l} \hline
    Technique          & Memory (MB)   & Time (secs)   & Accuracy       \\ \hline
    \multicolumn{4}{c}{\textbf{CIFAR-10}} \\ \hline            
    DARTS             & 21593$\pm$0   & 22924$\pm$154 & 96.79$\pm$0.20 \\
    $\beta$-DARTS       & 21593$\pm$0   & 23166$\pm$93  & 96.84$\pm$0.23   \\
    PC-DARTS          & 6438$\pm$0    & 19597$\pm$197 & 95.32$\pm$0.50 \\
    ETENAS            & 363$\pm$0     & 6159$\pm$30   & 96.11$\pm$0.32 \\
    \hline
    ETENAS-PC         & 379$\pm$0     & 5415$\pm$51   & 96.14$\pm$0.60 \\ 
    DARTS-RND  & 11190$\pm$978 & 12240$\pm$1058 & 96.63$\pm$0.07 \\
    \textbf{Zero-One-Shot (ours)}     & 4124$\pm$91   & 15175$\pm$332 & 96.86$\pm$0.14 \\
    \textbf{Zero-One-Shot-RND (ours)} & 3838$\pm$239  & 12252$\pm$924 & 96.51$\pm$0.29 \\
    \hline
    \multicolumn{4}{c}{\textbf{CIFAR-100}} \\ \hline

    DARTS           & 21594$\pm$0   & 22727$\pm$588  & 75.86$\pm$1.12 \\
    $\beta$-DARTS   & 21594$\pm$0   & 22937$\pm$372   & 80.01$\pm$0.47 \\
    PC-DARTS        & 6438$\pm$0    & 19619$\pm$199  & 77.41$\pm$1.97 \\
    ETENAS           & 364$\pm$0     & 6134$\pm$28    & 78.52$\pm$1.27 \\
    \hline
    ETENAS-PC         & 380$\pm$0     & 5456$\pm$18    & 79.10$\pm$1.16 \\
    DARTS-RND          & 12280$\pm$780 & 13152$\pm$1655 & 79.58$\pm$1.10  \\
    \textbf{Zero-One-Shot (ours)}     & 4084$\pm$43   & 14944$\pm$188  & 80.46$\pm$0.19 \\
    \textbf{Zero-One-Shot-RND (ours)} & 3933$\pm$178  & 12263$\pm$463  & 80.35$\pm$0.34 \\  
    \hline
    \end{tabular}
\end{table}

\subsection{Results}

\begin{figure}
    \centering
    \begin{subfigure}{0.49\textwidth}
        \includegraphics[width=\linewidth]{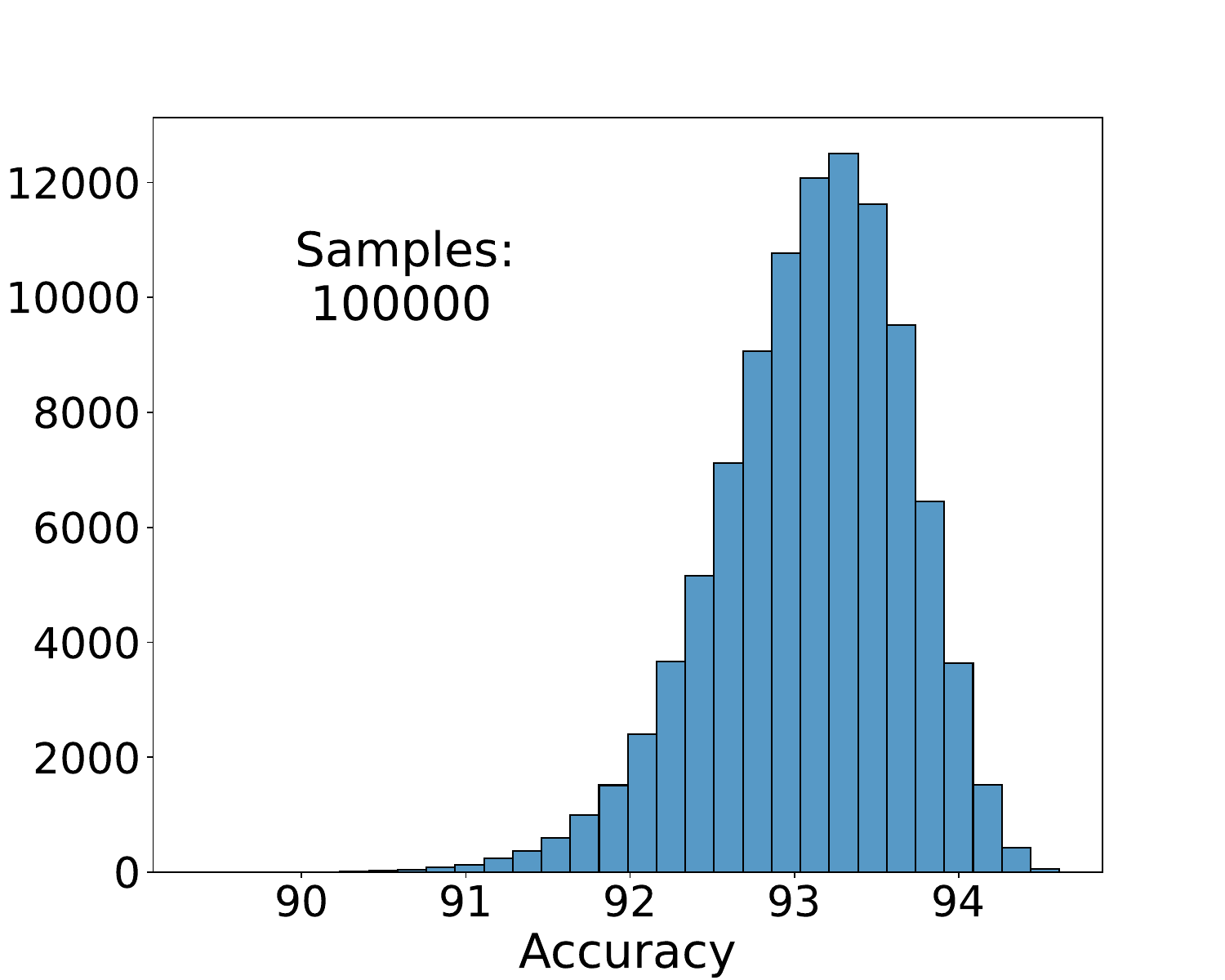}
        \caption{All architectures}
    \end{subfigure}
    \begin{subfigure}{0.49\textwidth}
        \includegraphics[width=\linewidth]{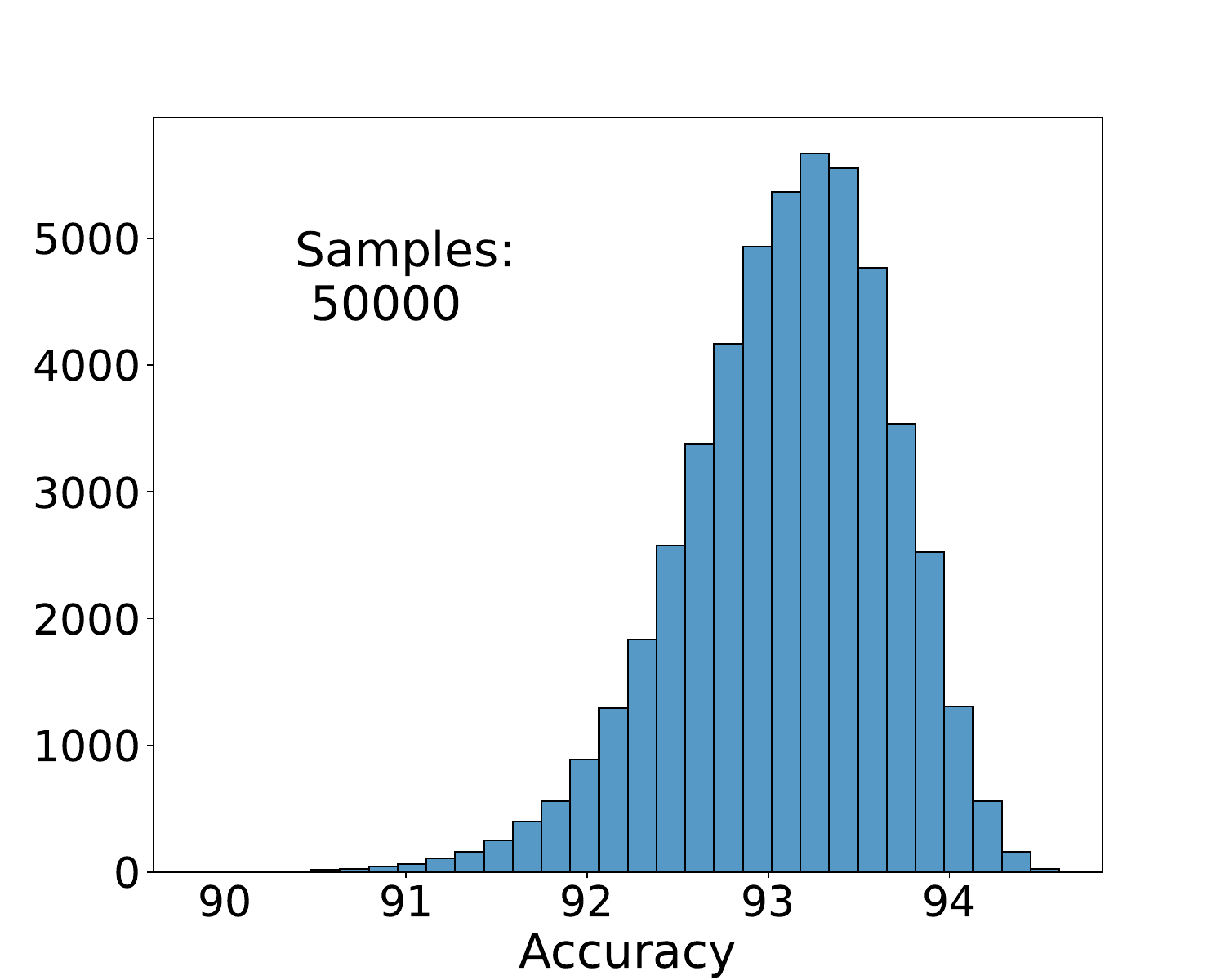}
        \caption{Half architectures masked}
    \end{subfigure}
    \caption{We sampled 100000 architectures from DARTS (has more than $10^{18}$ architectures) and obtained an approximation of their accuracies on CIFAR-10 using surrogate models of NAS-Bench-301~\citep{zela2022surrogate}. The distribution of  architectures (on the left) is heavily skewed towards top-performing architectures. We randomly removed half of the architectures (on the right) to demonstrate the minimal effect on the distribution.}
    \label{fig:nasbench301_distribution}
\end{figure}

We present the main results in Table~\ref{tab:memory_results}. 
We added baseline techniques for reference.
We use a naming convention of adding an ``-RND'' suffix for techniques that are utilising random pruning masks.

\textbf{Does pruning mask reduce memory requirements efficiently?}
Applying a pruning mask with $\xi = 0.5$ to the baseline DARTS reduces both memory requirements and search time by approximately half of the original values (see DARTS-RND).
When combined with additional memory reduction techniques, this approach enables a search within the same architectural space using only a fraction of the original memory requirements.
Our Zero-One-Shot requires only 19\% of the DARTS memory (36\% less than PC-DARTS).
It is 34\% faster than DARTS (23\% faster than PC-DARTS).
When the mask is randomly generated, our approach becomes 47\% faster than DARTS (37\% faster than PC-DARTS), while demonstrating higher performance than the baselines.

\textbf{How does memory reduction affect evaluation accuracy?}
In our experiments, applying a pruning mask did not result in any notable performance degradation.
Furthermore, reducing the search space generally led to improved accuracy and greater stability, even when using a randomly generated mask.
We hypothesise that DARTS is more stable on the smaller search space; however, the main source of stability might be $\beta$-regularisation.
Our observations also indicate that using a Zero-Shot ranking function to prune the search space yields a slightly higher evaluation accuracy compared to random pruning.

\textbf{Why does the application of random pruning masks not lead to significant accuracy deterioration?}
Zero-Shot architecture search is known for its strong capability to distinguish high-performing architectures from low-performing ones~\citep{ning2021evaluating}.
Applying a random pruning mask raises the concern that top-tier architectures might be excluded.
This can be a considerable issue on a small search space.
However, in the case of the DARTS search space, it is mitigated by the vast size of more than $10^{18}$ architectures.
Furthermore, the performance distribution within DARTS is heavily skewed towards high-performing architectures.
Consequently, the probability of all suitable candidates being simultaneously masked is negligible.
We illustrate this point in Figure~\ref{fig:nasbench301_distribution} with $10^5$ architectures randomly sampled from DARTS (less than $10^{-10}\%$ of all DARTS architectures).
Due to resource constraints, training all sampled architectures is infeasible; instead, we used NAS-Bench-301’s surrogate models~\citep{zela2022surrogate} to predict their accuracy on CIFAR-10.
The results indicate the skew towards top-performing architectures, with random pruning of half of the architectures having close to no effect on the distribution shape.

\section{Conclusion}

This paper introduces a streamlined framework for search time and memory reduction.
It is fully compatible with popular One-Shot architecture search methods commonly employed in practical applications.
We designed our framework to integrate seamlessly with any typical DARTS-like setup.
At its core, our approach prunes the architecture search space prior to applying One-Shot architecture search.
We introduce two methods for pruning: a faster, albeit less precise, random pruning and a more accurate, though slower, Zero-Shot-based approach.
In particular, our framework reduces the usage of DARTS memory by 81\% while identifying architectures with better performance.

The main limitation of search space pruning is the potential exclusion of top-tier architectures due to pruning.
On the search space, where the number of top-tier architectures is smaller than on DARTS, there is a chance of inadequately pruning top-tier architectures, thus diminishing the performance.
In general, we are interested in applying NAS to settings with very large search spaces, so the impact of pruning is likely to be negligible.

Our findings deliver an important message that large search spaces contain significant redundancy that can be eliminated. Zero-shot NAS can play a useful role in guiding the reduction, but even random pruning is effective.
In this paper, we demonstrate the concept by pruning half of the search space while retaining or improving the accuracy.
Furthermore, we establish a connection between Zero-Shot and One-Shot NAS, showcasing how their combined use can compensate for each other's limitations and leverage their strengths.
We believe that further integration of Zero-Shot and One-Shot methodologies could lead to more stable and memory-efficient architecture search techniques.



\bibliography{biblio}

\begin{thebibliography}{}

\bibitem[Abdelfattah et~al., 2021]{abdelfattah2021zerocost}
Abdelfattah, M.~S., Mehrotra, A., Dudziak, {\L}., and Lane, N.~D. (2021).
\newblock {Zero-Cost Proxies for Lightweight {NAS}}.
\newblock In {\em Proc. Int. Conf. Learn. Representations}.

\bibitem[Cai et~al., 2023]{cai2023epc}
Cai, Z., Chen, L., and Liu, H.-L. (2023).
\newblock {EPC-DARTS}: Efficient partial channel connection for differentiable architecture search.
\newblock {\em Neural Netw.}, 166:344--353.

\bibitem[Cavagnero et~al., 2023a]{cavagnero2023freerea}
Cavagnero, N., Robbiano, L., Caputo, B., and Averta, G. (2023a).
\newblock {FreeREA}: Training-free evolution-based architecture search.
\newblock In {\em Proc. IEEE Winter Conf. Appl. Comput. Vis.}

\bibitem[Cavagnero et~al., 2023b]{cavagnero2023entropic}
Cavagnero, N., Robbiano, L., Pistilli, F., Caputo, B., and Averta, G. (2023b).
\newblock {Entropic Score metric}: Decoupling topology and size in training-free {NAS}.
\newblock In {\em Proc. Int. Conf. Comput. Vis.}

\bibitem[Chen et~al., 2021]{chen2021tenas}
Chen, W., Gong, X., and Wang, Z. (2021).
\newblock Neural architecture search on imagenet in four {GPU} hours: A theoretically inspired perspective.
\newblock In {\em Proc. Int. Conf. Learn. Representations}.

\bibitem[Chen et~al., 2019]{chen2019progressive}
Chen, X., Xie, L., Wu, J., and Tian, Q. (2019).
\newblock Progressive differentiable architecture search: Bridging the depth gap between search and evaluation.
\newblock In {\em Proc. Int. Conf. Comput. Vis.}, pages 1294--1303.

\bibitem[Chu et~al., 2021]{chu2021darts}
Chu, X., Wang, X., Zhang, B., Lu, S., Wei, X., and Yan, J. (2021).
\newblock {DARTS-}: Robustly stepping out of performance collapse without indicators.
\newblock In {\em Proc. Int. Conf. Learn. Representations}.

\bibitem[Chu et~al., 2020]{chu2020fair}
Chu, X., Zhou, T., Zhang, B., and Li, J. (2020).
\newblock {FairDARTS}: Eliminating unfair advantages in differentiable architecture search.
\newblock In {\em Proc. Eur. Conf. Comput. Vis.}, pages 465--480.

\bibitem[DeVries, 2017]{devries2017cutout}
DeVries, T. (2017).
\newblock Improved regularization of convolutional neural networks with cutout.
\newblock {\em arXiv preprint arXiv:1708.04552}.

\bibitem[Dong and Yang, 2019]{dong2019searching}
Dong, X. and Yang, Y. (2019).
\newblock Searching for a robust neural architecture in four gpu hours.
\newblock In {\em Proc. IEEE Conf. Comput. Vis. \& Pattern Recognit.}, pages 1761--1770.

\bibitem[Elsken et~al., 2019]{elsken2019neural}
Elsken, T., Metzen, J.~H., and Hutter, F. (2019).
\newblock Neural architecture search: A survey.
\newblock {\em J. Mach. Learn. Res.}, 20(55):1--21.

\bibitem[Hu et~al., 2024]{hu2024ldarts}
Hu, L., Wang, Z., Li, H., Wu, P., Mao, J., and Zeng, N. (2024).
\newblock {$l$-DARTS}: Light-weight differentiable architecture search with robustness enhancement strategy.
\newblock {\em Knowledge-Based Syst.}, 288:111466.

\bibitem[Kang et~al., 2023]{kang2023neural}
Kang, J.-S., Kang, J., Kim, J.-J., Jeon, K.-W., Chung, H.-J., and Park, B.-H. (2023).
\newblock Neural architecture search survey: A computer vision perspective.
\newblock {\em Sensors}, 23(3):1713.

\bibitem[Kingma and Ba, 2014]{kingma2014adam}
Kingma, D.~P. and Ba, J. (2014).
\newblock {Adam: A Method for Stochastic Optimization}.
\newblock {\em arXiv preprint arXiv:1412.6980}.

\bibitem[Krizhevsky, 2009]{krizhevsky2009learning}
Krizhevsky, A. (2009).
\newblock Learning multiple layers of features from tiny images.
\newblock Technical report.

\bibitem[Li et~al., 2022]{li2022zico}
Li, G., Yang, Y., Bhardwaj, K., and Marculescu, R. (2022).
\newblock {ZiCo}: Zero-shot nas via inverse coefficient of variation on gradients.
\newblock In {\em Proc. Int. Conf. Learn. Representations}.

\bibitem[Lin et~al., 2021]{lin2021zen}
Lin, M., Wang, P., Sun, Z., Chen, H., Sun, X., Qian, Q., Li, H., and Jin, R. (2021).
\newblock {Zen-NAS}: A zero-shot nas for high-performance deep image recognition.
\newblock In {\em Proc. Int. Conf. Comput. Vis.}

\bibitem[Liu et~al., 2018]{liu2018darts}
Liu, H., Simonyan, K., and Yang, Y. (2018).
\newblock {DARTS}: Differentiable architecture search.
\newblock In {\em Proc. Int. Conf. Learn. Representations}.

\bibitem[Mellor et~al., 2020]{mellor2020neural}
Mellor, J., Turner, J., Storkey, A., and Crowley, E.~J. (2020).
\newblock Neural architecture search without training.
\newblock {\em arXiv preprint arXiv:2006.04647v1}.

\bibitem[Mellor et~al., 2021]{mellor2021neural}
Mellor, J., Turner, J., Storkey, A., and Crowley, E.~J. (2021).
\newblock Neural architecture search without training.
\newblock In {\em Proc. Int. Conf. Mach. Learn.}

\bibitem[Mok et~al., 2022]{mok2022demystifying}
Mok, J., Na, B., Kim, J.-H., Han, D., and Yoon, S. (2022).
\newblock Demystifying the neural tangent kernel from a practical perspective: Can it be trusted for neural architecture search without training?
\newblock In {\em Proc. IEEE Conf. Comput. Vis. \& Pattern Recognit.}

\bibitem[Movahedi et~al., 2023]{movahedi2023lambdadarts}
Movahedi, S., Adabinejad, M., Imani, A., Keshavarz, A., Dehghani, M., Shakery, A., and Araabi, B.~N. (2023).
\newblock $\lambda$-{DARTS}: Mitigating performance collapse by harmonizing operation selection among cells.
\newblock In {\em Proc. Int. Conf. Learn. Representations}.

\bibitem[Ning et~al., 2021]{ning2021evaluating}
Ning, X., Tang, C., Li, W., Zhou, Z., Liang, S., Yang, H., and Wang, Y. (2021).
\newblock Evaluating efficient performance estimators of neural architectures.
\newblock In {\em Adv. Neural Inf. Process. Syst.}

\bibitem[Park et~al., 2020]{park2020nngp}
Park, D.~S., Lee, J., Peng, D., Cao, Y., and Sohl-Dickstein, J. (2020).
\newblock Towards {NNGP}-guided neural architecture search.
\newblock {\em arXiv preprint arXiv:2011.06006}.

\bibitem[Rumiantsev and Coates, 2023]{rumiantsev2023performing}
Rumiantsev, P. and Coates, M. (2023).
\newblock Performing neural architecture search without gradients.
\newblock In {\em Proc. IEEE Int. Conf. Acoust. Speech, Signal Proc.}

\bibitem[Sun et~al., 2023]{sun2023unleashing}
Sun, Z., Sun, Y., Yang, L., Lu, S., Mei, J., Zhao, W., and Hu, Y. (2023).
\newblock Unleashing the power of gradient signal-to-noise ratio for zero-shot {NAS}.
\newblock In {\em Proc. IEEE Conf. Comput. Vis.}, pages 5763--5773.

\bibitem[Wang et~al., 2023]{wang2023fp}
Wang, W., Zhang, X., Cui, H., Yin, H., and Zhang, Y. (2023).
\newblock {FP-DARTS}: Fast parallel differentiable neural architecture search for image classification.
\newblock {\em Pattern Recognit.}, 136:109193.

\bibitem[Xu et~al., 2021]{xu2021knas}
Xu, J., Zhao, L., Lin, J., Gao, R., Sun, X., and Yang, H. (2021).
\newblock {KNAS}: green neural architecture search.
\newblock In {\em Proc. Int. Conf. Mach. Learn.}

\bibitem[Xu et~al., 2019]{xu2019pcdarts}
Xu, Y., Xie, L., Zhang, X., Chen, X., Qi, G.-J., Tian, Q., and Xiong, H. (2019).
\newblock {PC-DARTS}: Partial channel connections for memory-efficient architecture search.
\newblock In {\em Proc. Int. Conf. Learn. Representations}.

\bibitem[Yang et~al., 2023]{yang2023sweet}
Yang, L., Fu, Y., Lu, S., Sun, Z., Mei, J., Zhao, W., and Hu, Y. (2023).
\newblock {Sweet Gradient} matters: Designing consistent and efficient estimator for zero-shot architecture search.
\newblock {\em Neural Netw.}, 168:237--255.

\bibitem[Ye et~al., 2022]{ye2022bdarts}
Ye, P., Li, B., Li, Y., Chen, T., Fan, J., and Ouyang, W. (2022).
\newblock $\beta$-{DARTS}: Beta-decay regularization for differentiable architecture search.
\newblock In {\em Proc. IEEE Conf. Comput. Vis. \& Pattern Recognit.}, pages 10874--10883.

\bibitem[Zela et~al., 2020]{zela2020Understanding}
Zela, A., Elsken, T., Saikia, T., Marrakchi, Y., Brox, T., and Hutter, F. (2020).
\newblock Understanding and robustifying differentiable architecture search.
\newblock In {\em Proc. Int. Conf. Learn. Representations}.

\bibitem[Zela et~al., 2022]{zela2022surrogate}
Zela, A., Siems, J.~N., Zimmer, L., Lukasik, J., Keuper, M., and Hutter, F. (2022).
\newblock Surrogate {NAS} benchmarks: Going beyond the limited search spaces of tabular {NAS} benchmarks.
\newblock In {\em Proc. Int. Conf. Learn. Representations}.

\bibitem[Zoph et~al., 2018]{zoph2018learning}
Zoph, B., Vasudevan, V., Shlens, J., and Le, Q.~V. (2018).
\newblock Learning transferable architectures for scalable image recognition.
\newblock In {\em Proc. IEEE Conf. Comput. Vis. \& Pattern Recognit.}, pages 8697--8710.

\end{thebibliography}




\newpage
\section*{Submission Checklist}


\begin{enumerate}
\item For all authors\dots
  \begin{enumerate}
  \item Do the main claims made in the abstract and introduction accurately reflect the paper's contributions and scope?
    \answerYes{}
  \item Did you describe the limitations of your work?
    \answerYes{}
  \item Did you discuss any potential negative societal impacts of your work?
    \answerNA{}
  \item Did you read the ethics review guidelines and ensure that your paper
    conforms to them? (see \url{https://2022.automl.cc/ethics-accessibility/})
    \answerYes{}
  \end{enumerate}
\item If you ran experiments\dots
  \begin{enumerate}
  \item Did you use the same evaluation protocol for all methods being compared (e.g.,
    same benchmarks, data (sub)sets, available resources, etc.)?
    \answerYes{}
  \item Did you specify all the necessary details of your evaluation (e.g., data splits,
    pre-processing, search spaces, hyperparameter tuning details and results, etc.)?
    \answerYes{}
  \item Did you repeat your experiments (e.g., across multiple random seeds or
    splits) to account for the impact of randomness in your methods or data?
    \answerYes{}
  \item Did you report the uncertainty of your results (e.g., the standard error
    across random seeds or splits)?
    \answerYes{}
  \item Did you report the statistical significance of your results?
    \answerNo{Our primary metric is Memory Consumption. The difference with main baseline (PC-DARTS) in beyong 25 standart deviations.}
  \item Did you use enough repetitions, datasets, and/or benchmarks to support
    your claims?
    \answerYes{}
  \item Did you compare performance over time and describe how you selected the
    maximum runtime?
    \answerYes{}
  \item Did you include the total amount of compute and the type of resources
    used (e.g., type of \textsc{gpu}s, internal cluster, or cloud provider)?
    \answerYes{}
  \item Did you run ablation studies to assess the impact of different
    components of your approach?
    \answerYes{}
  \end{enumerate}
\item With respect to the code used to obtain your results\dots
  \begin{enumerate}
\item Did you include the code, data, and instructions needed to reproduce the
    main experimental results, including all dependencies (e.g.,
    \texttt{requirements.txt} with explicit versions), random seeds, an instructive
    \texttt{README} with installation instructions, and execution commands
    (either in the supplemental material or as a \textsc{url})?
    \answerYes{}
  \item Did you include a minimal example to replicate results on a small subset
    of the experiments or on toy data?
    \answerNo{Small example in unfeasible, any example it still requires the training of the identified architecture.}
  \item Did you ensure sufficient code quality and documentation so that someone else
    can execute and understand your code?
    \answerYes{}
  \item Did you include the raw results of running your experiments with the given
    code, data, and instructions?
    \answerYes{We have included training and search logs.}
  \item Did you include the code, additional data, and instructions needed to generate
    the figures and tables in your paper based on the raw results?
    \answerYes{}
  \end{enumerate}
\item If you used existing assets (e.g., code, data, models)\dots
  \begin{enumerate}
  \item Did you cite the creators of used assets?
    \answerYes{}
  \item Did you discuss whether and how consent was obtained from people whose
    data you're using/curating if the license requires it?
    \answerNA{}
  \item Did you discuss whether the data you are using/curating contains
    personally identifiable information or offensive content?
    \answerNA{}
  \end{enumerate}
\item If you created/released new assets (e.g., code, data, models)\dots
  \begin{enumerate}
    \item Did you mention the license of the new assets (e.g., as part of your
    code submission)?
    \answerYes{The license is included into the released code}
    \item Did you include the new assets either in the supplemental material or as
    a \textsc{url} (to, e.g., GitHub or Hugging Face)?
    \answerYes{The code is included as a suplimentary material}
  \end{enumerate}
\item If you used crowdsourcing or conducted research with human subjects\dots
  \begin{enumerate}
  \item Did you include the full text of instructions given to participants and
    screenshots, if applicable?
    \answerNA{}
  \item Did you describe any potential participant risks, with links to
    institutional review board (\textsc{irb}) approvals, if applicable?
    \answerNA{}
  \item Did you include the estimated hourly wage paid to participants and the
    total amount spent on participant compensation?
    \answerNA{}
  \end{enumerate}
\item If you included theoretical results\dots
  \begin{enumerate}
  \item Did you state the full set of assumptions of all theoretical results?
    \answerNA{}
  \item Did you include complete proofs of all theoretical results?
    \answerNA{}
  \end{enumerate}
\end{enumerate}

\newpage
\appendix



\end{document}